%% file: acl_latex.tex
\pdfoutput=1

\documentclass[11pt]{article}

\usepackage[preprint]{acl}

\usepackage{times}
\usepackage{latexsym}
\usepackage{CJKutf8}
\usepackage[T1]{fontenc}

\usepackage[utf8]{inputenc}

\usepackage{microtype}

\usepackage{inconsolata}

\usepackage{amsmath}
\usepackage{amsfonts}
\usepackage{amssymb}
\usepackage{subfigure}
\usepackage{cases}
\usepackage{dsfont}
\usepackage{bm}
\usepackage{bbm}
\usepackage{graphicx}
\usepackage{color}
\usepackage{multicol}
\usepackage{multirow}
\usepackage{wrapfig,lipsum,booktabs}
\usepackage[ruled,vlined,linesnumbered]{algorithm2e}
\usepackage{pgfplots}
\pgfplotsset{compat=1.12} 
\usepackage{filecontents}
\usepackage{tikz}
\usepackage{xcolor}
\usepackage{colortbl}
\usepackage{xspace}
\usepackage{makecell}
\usepackage{soul}
\usepackage{subcaption}
\usetikzlibrary{calc}
\usepgfplotslibrary{groupplots}
\usetikzlibrary{angles,quotes} 
\usetikzlibrary{shapes,arrows}
\usetikzlibrary{backgrounds}
\usetikzlibrary{matrix}
\usetikzlibrary{patterns}
\usepackage{tikz-3dplot}
\usepackage{hyperref}
\usepackage{cleveref}
\usepackage{paralist}
\usepackage{cancel}
\usepackage{todonotes}
\usepackage{tabu}
\usepackage{rotating}
\usepackage{etoolbox}
\usepackage{adjustbox}
\usepackage{enumerate}
\usepackage{enumitem}
\setitemize{itemsep=0pt,topsep=0pt,parsep=0pt,partopsep=0pt}
\setenumerate{itemsep=0pt,topsep=0pt,parsep=0pt,partopsep=0pt}
\usepackage{pifont}
\usepackage{cancel}
\usepackage{lipsum}
\usepackage{listings,lstautogobble}
\usepackage{fancyvrb}
\usepackage{fvextra}
\usepackage{caption}
\usepackage{arydshln}
\usepackage{float}
\usepackage{pmboxdraw} 

\newcommand{\fullname}[1]{\textbf{#1}\xspace}
\newcommand{\oursfull}{\fullname{Inference-Time Cross-Lingual Intervention}}

\newcommand{\model}[1]{\textsc{#1}\xspace}

\newcommand{\baseline}{\model{Baseline}}
\newcommand{\bloomz}{\model{BLOOMZ-7b1-mt}}
\newcommand{\llama}{\model{Llama2-7B-chat}}
\newcommand{\llamabase}{\model{Llama2-7B}}
\newcommand{\llamathree}{\model{Llama3-8B-instruct}}
\newcommand{\mistral}{\model{Mistral-7b-instruct}}
\newcommand{\falcon}{\model{Falcon-7b-instruct}}

\newcommand{\iti}{\model{ITI}}
\newcommand{\caa}{\model{CAA}}
\newcommand{\mtgoogle}{\model{MT-Google}}
\newcommand{\mtllm}{\model{MT-LLM}}
\newcommand{\sft}{\model{SFT}}
\newcommand{\mathllama}{\model{MathOctopus}}

\newcommand{\ours}{\model{INCLINE}}
\newcommand{\ourshidden}{\model{INCLINE-hidden}}
\newcommand{\oursattn}{\model{INCLINE-attn}}

\newcommand{\oursffn}{\model{INCLINE-ffn}}
\newcommand{\oursemb}{\model{INCLINE-emb}}
\newcommand{\oursfdev}{\model{INCLINE-Fdev}}

\newcommand{\dataset}[1]{\texttt{#1}\xspace}
\newcommand{\xcopa}{\dataset{XCOPA}}
\newcommand{\xstorycloze}{\dataset{XStoryCloze}}
\newcommand{\xwinograd}{\dataset{XWinograd}}
\newcommand{\xcsqa}{\dataset{XCSQA}}
\newcommand{\xnli}{\dataset{XNLI}}
\newcommand{\mzsre}{\dataset{MZsRE}}
\newcommand{\flores}{\dataset{Flores}}
\newcommand{\mgsm}{\dataset{MGSM}}
\newcommand{\wmt}{\dataset{WMT23}}

\newcommand{\vx}{\pmb{x}}

\newcommand{\vh}{\pmb{h}}

\newcommand{\vq}{\pmb{q}}

\newcommand{\avgall}{\mu_{\textsc{all}}}
\newcommand{\avgseen}{\mu_{\textsc{seen}}}
\newcommand{\avgunseen}{\mu_{\textsc{unseen}}}

\DeclareMathOperator*{\argmin}{argmin}

\title{Bridging the Language Gaps in Large Language Models with Inference-Time Cross-Lingual Intervention}

\author{%
Weixuan Wang\textsuperscript{1$\dagger$} \quad Minghao Wu\textsuperscript{2$\dagger$} \quad Barry Haddow\textsuperscript{1} \quad Alexandra Birch\textsuperscript{1} \\[1ex]
\textsuperscript{1}School of Informatics, University of Edinburgh \\
\textsuperscript{2}Monash University \\
\texttt{\{weixuan.wang, bhaddow, a.birch\}@ed.ac.uk} \\
\texttt{minghao.wu@monash.edu} 
}

\begin{document}

\renewcommand{\tableautorefname}{Table}
\renewcommand{\sectionautorefname}{Section}
\renewcommand{\subsectionautorefname}{Section}
\renewcommand{\subsubsectionautorefname}{Section}
\renewcommand{\figureautorefname}{Figure}
\newcommand{\subfigureautorefname}{Figure}
\renewcommand{\algorithmautorefname}{Algorithm}
\newcommand{\linenoautorefname}{Line}
\renewcommand{\appendixname}{Appendix}

\maketitle

\begingroup
\renewcommand\thefootnote{}\footnotetext{
\textsuperscript{$\dagger$}~Equal contribution.
}
\endgroup

\begin{abstract}

Large Language Models (LLMs) have shown remarkable capabilities in natural language processing but exhibit significant performance gaps among different languages. Most existing approaches to address these disparities rely on pretraining or fine-tuning, which are resource-intensive. To overcome these limitations without incurring significant costs, we propose \textbf{\oursfull (\ours)}, a novel framework that enhances LLM performance on low-performing (source) languages by aligning their internal representations with those of high-performing (target) languages during inference. \ours initially learns alignment matrices using parallel sentences from source and target languages through a Least-Squares optimization, and then applies these matrices during inference to transform the low-performing language representations toward the high-performing language space. Extensive experiments on nine benchmarks with five LLMs demonstrate that \ours significantly improves performance across diverse tasks and languages, compared to recent strong baselines. Our analysis demonstrates that \ours is highly cost-effective and applicable to a wide range of applications. In addition, we release the code to foster research along this line.\footnote{https://github.com/weixuan-wang123/INCLINE}

\end{abstract}

\input{1_introduction}

\input{2_related_work}

\input{3_method}
\input{4_experiments}
\input{5_analysis_and_discussion}

\input{6_discussion}

\input{7_conclusion}

\input{9_limitations}

\bibliography{custom,anthology}

\clearpage
\appendix

\input{8_appendix}

\end{document}

%% file: 1_introduction.tex
\section{Introduction}
\label{sec:introduction}

\begin{figure}[t]
\centering          
\subfigure[Before intervention]{\label{fig:projection-ori}\includegraphics[scale=0.37]{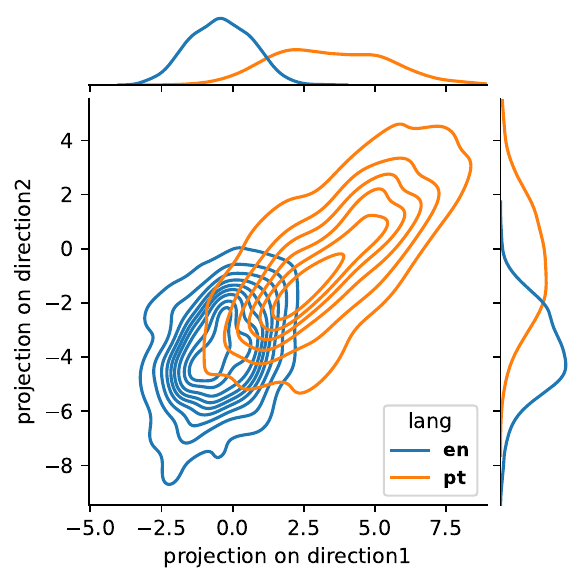}} 
\subfigure[After intervention]{\label{fig:projection-incline}\includegraphics[scale=0.37]{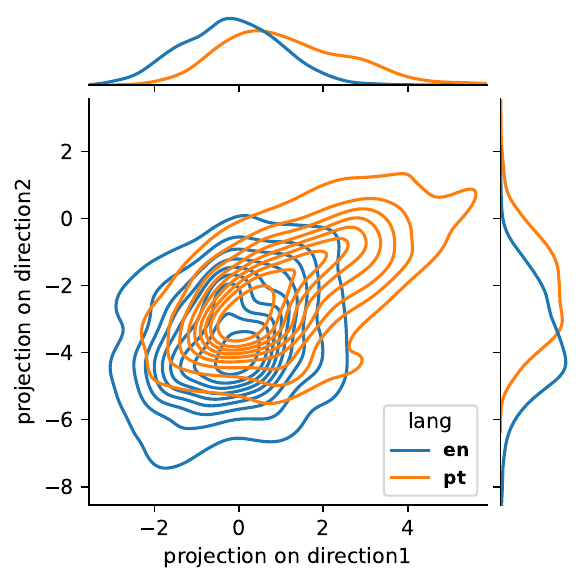}} 
\caption{Bivariate kernel density estimation plots displaying the representations (hidden states of the last token) from 100 random examples in English (blue) and their Portuguese translations (orange) from \xcopa \citep{xcopa}. After intervention using \ours, the Portuguese representations are aligned closer to the English representations.}
\label{fig:kernel_project}
\end{figure}

Large Language Models (LLMs) have achieved remarkable success across a variety of natural language processing tasks, demonstrating strong capabilities in language understanding and generation \citep{gpt4,llama3,gemma,claudesonnet,gpt4o,o1}. However, despite these advancements, most state-of-the-art LLMs remain predominantly English-centric, exhibiting significant performance gaps among different languages \citep{gap1,gap}, which can adversely affect user experience and potentially exclude large portions of the global population from accessing advanced AI services \citep{cultral0,cultral1}.

Addressing the performance gaps across languages is highly challenging. Recent approaches are mostly data-driven, such as multilingual supervised fine-tuning or continued pre-training \citep{aya,continue,continue1}. However, collecting and annotating large-scale datasets for numerous underrepresented languages is both time-consuming and resource-intensive \citep{okapi}. Furthermore, training LLMs on multilingual data requires substantial computational resources, limiting their practicality for widespread applications, especially in resource-constrained settings \citep{bloomz,bactrian}. Given these limitations, a natural question arises: \textit{How can we bridge the performance gaps between high-performing and low-performing languages without incurring prohibitive costs?}

Inspired by \citet{wordtrans} showing that word embeddings in different languages can be aligned to a shared representation space through learned rotations for word translation, we propose \textbf{\oursfull (\ours)}. This novel framework utilizes a group of learned alignment matrices that  transform the representations (e.g., hidden states) of a low-performing (source) language into those of a high-performing (target) language during inference. Our framework comprises two main steps. First, we train the alignment matrices for each layer of LLM using parallel sentences from the source and target languages. The learning process is formulated as a Least-Squares optimization problem, where these alignment matrices are learned by minimizing the distance between the projected source language representations and their corresponding target language representations, without the need for extensive retraining or fine-tuning the LLM. Second, we apply the learned alignment matrices to transform the source language input representations into the target language representation space at each layer during inference. By integrating these steps, \ours leverages the rich representations learned from high-performing languages to enhance performance on downstream tasks involving low-performing languages. As shown in \autoref{fig:kernel_project}, \ours effectively aligns the input representations in Portuguese to their parallel representations in English.

In this study, we conduct extensive experiments to validate the effectiveness of \ours on nine widely used benchmarks using five LLMs. Our results demonstrate that aligning internal representations using \ours significantly improves performance on diverse tasks among languages. 

Our contributions are summarized as follows:
\begin{itemize}
    \item We propose \ours, a cross-lingual intervention approach that enhances LLMs by transforming source language representations into a target language representation space during inference without requiring additional training of LLMs (see \autoref{sec:method}).

    \item We conduct extensive evaluations across five discriminative tasks and four generative tasks, covering 21 languages. Our experimental results show that \ours significantly improves model performance, boosting average accuracy by up to +4.96 compared to strong baselines (see \autoref{sec:experiments}).

    \item Our detailed analysis indicates that \ours is highly cost-effective, as it requires minimal computational resources while delivering substantial performance improvements (see \autoref{sec:analysis}). Moreover, we demonstrate that \ours is effective with regard to LLM backbones, model sizes, and in-context learning, underscoring its general applicability and potential for broader use in enhancing LLMs for underrepresented languages (see \autoref{sec:discussions}).
\end{itemize}

\begin{figure*}[t]
    \centering
    \includegraphics[scale=0.45]{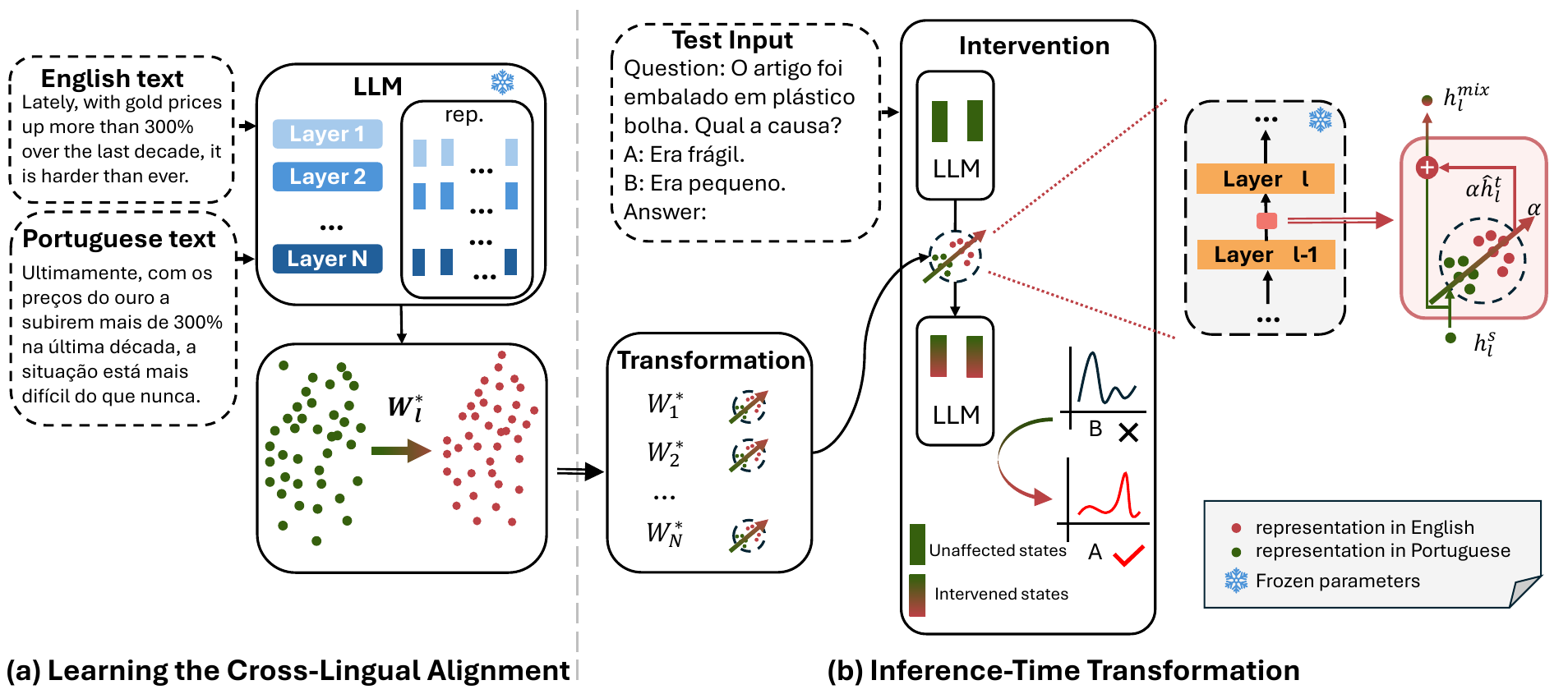}
    \caption{Framework of \ours. \ours involves two steps: (a) Learning the Cross-Lingual Alignment: sentence representations from a parallel dataset are used to train alignment matrices that map source (Portuguese) representations to the target (English) representations. (b) Inference-Time Transformation: this step adapts the source representations from downstream tasks into the target representation space using the alignment matrices.}
    \label{fig:framework}
\end{figure*}

%% file: 2_related_work.tex
\section{Related Work}
\label{sec:related}
\paragraph{Multilingual LLMs}
LLMs are pivotal in multilingual NLP tasks, typically leveraging external parallel datasets for training \citep{mt5,bloomz,palm}. For low-resource languages, data augmentation techniques generate parallel data by mining sentence pairs or translating monolingual text using machine translation tools \citep{back,contrastive0,translate0}. However, these methods heavily rely on robust parallel corpora. To reduce data costs, studies have shifted toward Parameter-Efficient Fine-Tuning (PEFT) techniques \citep{mad-x,bad-x,example,translationtask1} and cross-lingual embeddings mapping methods \citep{emb0,emb1,emb2}, which still demand considerable computational resources.

\paragraph{Multilingual Prompting}
There is a growing interest in methods that do not require parameter adjustments. Prompting techniques have emerged, utilizing LLMs with multilingual prompts \citep{prompt1,prompt0,prompt2,prompt3}. However, these strategies face challenges like poor translation quality and prompt framing interference \citep{reliability}. Additionally, their effectiveness varies by task, as recent research indicates that few-shot learning may not outperform zero-shot learning in translation tasks \citep{translationtask}.

\paragraph{Intervention}
To address these challenges, we explore inference-time intervention techniques as cost-effective and efficient alternatives to traditional fine-tuning. Prior research in style transfer \citep{style0,style1}, knowledge editing \citep{ke0}, and truthfulness shifting \citep{iti,caa} demonstrates the potential of linear probe-based interventions. However, these methods have been largely limited to monolingual contexts. 
Our goal is to design a novel cross-lingual inference-time intervention that effectively aligns representations across languages, aiming to improve performance across multiple languages.

%% file: 3_method.tex
\section{Methodology}
\label{sec:method}
In \autoref{fig:framework}, we illustrate the framework of \ours, which enhances LLMs through inference-time cross-lingual intervention. Our approach comprises two main steps:
\begin{itemize} 
\item \textbf{Learning the Cross-Lingual Alignment}: Using parallel corpora, we train alignment matrices for each layer to map ource language representations totarget language representations (see \autoref{sec:method_learn}).
\item \textbf{Inference-Time Transformation}: During inference, we utilize the learned alignment matrices to transform input representations from the source language into the target language representation space, thereby improving the LLM's performance on tasks in the source language (see \autoref{sec:method_infer}).
\end{itemize}
By minimizing the distance between the source language representations and their corresponding target language representations, we effectively reduce cross-lingual representation gaps and align representation spaces across languages.

\subsection{Learning the Cross-Lingual Alignment}
\label{sec:method_learn}

Inspired by \citet{embedding-linear} that align embeddings across languages with learned linear transformations, we aim to learn a cross-lingual alignment matrix $\bm{W}_{l}$ that aligns sentence representations from the source language to the target language at the $l$-th layer of LLM. Given a parallel dataset $D = \{(\vx_i^{\textrm{s}}, \vx_i^{\textrm{t}})\}_{i=1}^{N}$, where each $\vx_i^{\textrm{s}}$ is the $i$-th source sentence and $\vx_i^{\textrm{t}}$ is its corresponding translation in the target language. Both $ \vx_i^{\textrm{s}} $ and $ \vx_i^{\textrm{t}} $ are sequences of tokens. From these sequences, we extract sentence representations by taking the hidden state of the last token in each sequence, denoted as $\vh_{i,l}^{\textrm{s}} \in \mathbb{R}^d $ and $ \vh_{i,l}^{\textrm{t}} \in \mathbb{R}^d $ for the source and target sentence, respectively, where $ d $ is the dimensionality of the hidden states.

To minimize the difference between the projected source sentence representations and the target sentence representations, our objective can be defined as a Least-Squares optimization problem:
\begin{align}
    \bm{W}^*_{l} = \argmin_{\bm{W}_{l}} \sum_{i=1}^N \left\| \bm{W}_{l} \vh_{i,l}^{\textrm{s}} - \vh_{i,l}^{\textrm{t}} \right\|^2
\end{align}
This problem seeks the optimal $\bm{W}^*_{l}$ that aligns the source representations with the target representations by minimizing the distance between them. Hence, the closed-form solution to this optimization problem is:

\begin{align}
    \bm{W}^*_{l} = 
    \left( \sum_{i=1}^N  (\vh_{i,l}^{\textrm{s}})^\top \vh_{i,l}^{\textrm{s}}  \right)^{-1} 
    \left( \sum_{i=1}^N (\vh_{i,l}^{\textrm{s}})^\top \vh_{i,l}^{\textrm{t}}  \right) 
\end{align}
By applying the learned alignment matrix $\bm{W}^*_{l}$ to the source sentence representations, we effectively map them into the target language's representation space. This alignment reduces cross-lingual representation discrepancies, allowing the model to leverage knowledge from the target language to improve performance on tasks in the source language.

\subsection{Inference-Time Transformation}
\label{sec:method_infer}

With the learned alignment matrix $\bm{W}_l^*$, we can enhance the LLM's processing of source language inputs by transforming their representations to the target representation space during inference.

We denote the hidden state of the last token of the test input $\vq^{s}$ in the source language at the $l$-th layer of the LLM as $\vh^{\textrm{s}}_{q,l}$ and then project this source language representation into the target representation space using the alignment matrix $\bm{W}^*_{l}$:
\begin{align}
    \hat{\vh}^{\textrm{t}}_{q,l} = \bm{W}^*_{l} \vh^{\textrm{s}}_{q,l} 
\end{align}
To perform the cross-lingual intervention at the $l$-th layer using the intervention vector $\hat{\vh}^{\textrm{t}}_{q,l}$, we adjust the original hidden state in source language $\vh^{\textrm{s}}_{q,l}$ by blending it with the projected hidden state in target language $\hat{\vh}^{\textrm{t}}_{q,l}$. This adjustment is controlled by a hyperparameter $\alpha$, which balances the influence between the source and target hidden states:
\begin{align}
\label{eq:mix}
    \vh^{\textrm{mix}}_{q,l} = \vh^{\textrm{s}}_{q,l} + \alpha \hat{\vh}^{\textrm{t}}_{q,l}
\end{align}
Here, \autoref{eq:mix} represents a shift of representation of source language towards target language representation by a magnitude of $\alpha$ times.  

\paragraph{Decoding with Minimal Intervention}
In this work, we only conduct one single intervention on the last token of $\vq^{\textrm{s}}$ by replacing $\vh^{\textrm{s}}_{q,l}$ with $\vh^{\textrm{mix}}_{q,l}$ for the test input $\vq^{s}$ at the $l$-th layer of LLM. In such a way, we can effectively intervene the model output while preserve the features in the source language.

\paragraph{Comparison with \iti and \caa}
Recently, \iti \citep{iti} and \caa \citep{caa} have been proposed as interventions in the model behaviors by manipulating the selected attention heads and hidden states, respectively. \ours is distinct from \iti and \caa due to three primary differences. Firstly, \iti and \caa utilize a learned \textit{static} intervention vector to alter model behaviors, whereas \ours leverages a set of alignment matrices to \textit{dynamically} align input representations from the source language to the target language. Secondly, \iti and \caa apply the intervention vector across all token positions following the instruction, potentially causing excessive perturbation during inference. In contrast, \ours performs a single intervention solely on the last token of the input. Additionally, unlike \iti and \caa, which target on only a limited number of layers, \ours modifies the representations across all layers. These modifications enable the LLMs to comprehensively leverage their target language capabilities for multilingual prediction.

%% file: 4_experiments.tex
\section{Experiments}
\label{sec:experiments}

In this section, we introduce our experimental setup (\autoref{sec:experiments_setup}) and present our results in \autoref{sec:experiments_results}.

\subsection{Experimental Setup}
\label{sec:experiments_setup}

We present our evaluation tasks, model backbones, implementation details of \ours, and baselines in this section.

\paragraph{Evaluation Tasks}

We conduct extensive evaluations across nine diverse downstream tasks, categorized into two groups: 
\begin{itemize}
    \item \textbf{Discriminative Tasks}: \xcopa \citep{xcopa}, \xstorycloze  \citep{xstorycloze}, \xwinograd~\citep{xstorycloze}, \xcsqa~\citep{xcsqa}, \xnli~\citep{xnli};
    \item \textbf{Generative Tasks}: \mzsre~\citep{mzsre}, \flores~\citep{flores}, \wmt~\citep{wmt23}, \mgsm~\citep{mgsm}.
\end{itemize}
These tasks covers 21 languages including English (en), Arabic (ar), German (de), Greek (el), Spanish (es), Estonian (et), French (fr), Hindi (hi), Indonesian (id), Italian (it), Japanese (ja), Dutch (nl), Portuguese (pt), Russian (ru), Swahili (sw), Tamil (ta), Thai (th), Turkish (tr), Ukrainian (uk), Vietnamese (vi), and Chinese (zh). We include more details of these tasks in \autoref{sec:appendix-dataset}.

\input{tables/main-mc}

\paragraph{Model Backbones}
In this work, we mainly use \bloomz as our model backbone for all the baseline approaches, unless otherwise specified. To demonstrate the effectiveness of \ours across various model backbones, we include four additional LLMs: \llamathree~\citep{llama3}, \llama~\citep{llama}, \mistral~\citep{mistral}, \falcon~\citep{falcon}. We present these results in \autoref{sec:discussions}. For the \mgsm task, we employ the \mathllama \citep{mathoctopus},\footnote{https://huggingface.co/Mathoctopus/Parallel\_7B} a specialized model fine-tuned from \llamabase for mathematical reasoning tasks, as the backbone. 

\paragraph{\ours (Ours)}
In this work, we mainly focus on aligning the low-performing language (source) representations closer to the English (target) representations, as LLMs are predominantly English-centric. 
For training the alignment matrices between languages, we randomly sample 500 parallel sentence pairs for each language pair involving English and other languages. These pairs are sourced from the News Commentary v16 dataset \citep{wmt19}, and for languages not covered by this dataset, we use the CCAligned dataset \citep{ccalign}. 
Following \citet{caa}, the value of the $\alpha$ controlling the intervention strength is in the range from -1 to 1 and determined by the validation results for each language across tasks. We use one A100 GPU (40G) for all experiments. 

\paragraph{Baselines}
We compare \ours against several established techniques: (1) \textbf{\baseline} indicates the predictions given by the original \bloomz; (2) \textbf{\mtgoogle} utilizes \model{Google Translate} to translate non-English questions into English; (3) \textbf{\mtllm} leverages \bloomz to translate questions in non-English languages into English, employing the structured prompt template ``\texttt{\{Source Language\}: \{Inputs\} English:}''; (4) \textbf{\sft} represents the task-specific supervised fine-tuning (SFT) involving updating all parameters of the LLM on the English training set for each downstream task individually with the hyperparameters described in \autoref{sec:appendix-sft_hyperparameters} and evaluating the resulting model on the multilingual test sets; 
(5) \textbf{\iti} \citep{iti} is an intervention method that identifies attention heads with high linear probing accuracy for truthfulness and adjusts activations along these truth-correlated directions during inference. Originally used to shift models from generating false statements to truthful ones, we adapt it to encourage the generation of English text over non-English text.
(6) \textbf{\caa} \citep{caa} employs the mean difference in hidden states between positive and negative examples from additional training data as an intervention vector to adjust the model's behavior towards the desired direction. Initially designed for monolingual alignment-relevant tasks, we utilize it to shift the model's output from non-English to English.

\subsection{Results}
\label{sec:experiments_results}

\input{tables/main-gen}

In this section, we present our results on the discriminative tasks (\autoref{tab:main-mc}) and generative tasks (\autoref{tab:main-gen}). Furthermore, we also categorize the languages involved in the downstream tasks into two groups based on the training data of \bloomz: \textit{seen languages} (ar, es, fr, hi, id, pt, sw, ta, vi, and zh) and \textit{unseen languages} (de, el, et, it, ja, nl, ru, th, tr, and uk). The breakdown results are provided in \autoref{tab:overall} (see \autoref{sec:appendix-detail}).

\paragraph{\ours significantly improves discriminative task performance.} 
The experimental results in \autoref{tab:main-mc} clearly demonstrate the effectiveness of \ours. Although methods like \sft, \mtgoogle, and \mtllm achieve high performance, they come with substantial costs, including the need for extensive fine-tuning of LLMs and reliance on third-party tools. Activation intervention methods, such as \iti and \caa, offer a more cost-effective solution but yield only minimal improvements, indicating a potential inadequacy in capturing the complexities of multilingual tasks. In contrast, \ours provides significant performance gains by enhancing multilingual representation alignment at inference time without requiring extensive resources or dependencies. This results in a more efficient improvement in multilingual performance. For example, \ours increases the average accuracy by +4.96 on \xstorycloze. Additionally, it delivers improvements of +4.20 and +9.46 for seen and unseen languages, respectively. Moreover, \ours can further improve the performance of the task-specific \sft.

\paragraph{\ours significantly enhances generative task performance.} 
The experimental results presented in \autoref{tab:main-gen} suggest the effectiveness of \ours in enhancing performance across generative tasks. Unlike \iti and \caa, which show only marginal improvements similar to those observed in discriminative tasks, \ours appears to achieve substantial advancements. Notably, \iti seems to struggle significantly in machine translation tasks, such as \flores and \wmt, highlighting its limitations. Furthermore, \ours reportedly boosts accuracy in the \mgsm task by up to +3.50 across various languages. 
This finding suggests that, although the mathematical capabilities are independent from the languages, understanding the questions written in different languages still requires language-specific knowledge. \ours successfully transfers the LLMs' natural language understanding capabilities from English to other languages. It is important to note that \sft is not evaluated on generative tasks because there are no training sets associated with these tasks.

In summary, these results demonstrate that \ours offers a significant improvement in both discriminative and generative tasks by effectively aligning multilingual representations.

%% file: tables/main-mc.tex
\begin{table*}[t] \scriptsize
\centering
\setlength{\tabcolsep}{2.5pt}
\begin{tabular}{lccccccccccccccc}     
\toprule
& \multicolumn{3}{c}{\xcopa}      & \multicolumn{3}{c}{\xstorycloze}  & \multicolumn{3}{c}{\xwinograd}    & \multicolumn{3}{c}{\xcsqa}  & \multicolumn{3}{c}{\xnli}   \\   \cmidrule(rl){2-4} \cmidrule(rl){5-7} \cmidrule(rl){8-10} \cmidrule(rl){11-13} \cmidrule(rl){14-16} 
& $\avgall$    & $\avgseen$     & $\avgunseen$   & $\avgall$  &   $\avgseen$     & $\avgunseen$   & $\avgall$        &        $\avgseen$     & $\avgunseen$ & $\avgall$        & $\avgseen$     & $\avgunseen$  & $\avgall$         & $\avgseen$     & $\avgunseen$      \\ \midrule
\baseline & 61.62  & 69.00  & 52.40  & 74.96    & 77.83  & 57.78  & 57.05     & 59.71  & 53.06 & 47.35     & 55.31 & 34.62  & 46.48     & 50.04 & 41.48  \\
\mtgoogle & 73.31\rlap{\textsuperscript{$\dagger$}}     & 73.52     & 73.05\rlap{\textsuperscript{$\dagger$}}     & 76.63    & 76.05  & 80.08\rlap{\textsuperscript{$\dagger$}}  & 57.63     & 57.12  & 57.90\rlap{\textsuperscript{$\dagger$}} & 58.52\rlap{\textsuperscript{$\dagger$}}  & 54.84 & 64.40\rlap{\textsuperscript{$\dagger$}}     & 50.72  & 49.80 & 52.00 \\
\mtllm    & 59.84  & 67.16  & 50.70  & 79.41    & 82.23  & 62.48  & 43.02     & 41.67  & 45.04 & 30.73     & 35.38 & 23.30  & 43.64     & 47.83 & 37.77  \\
\multicolumn{16}{l}{\cellcolor{gray!15}\textbf{Intervention Methods}}   \\
\iti      & 60.91  & 67.56  & 52.60  & 76.38    & 79.33  & 58.70  & 48.24     & 58.37  & 33.06 & 46.32     & 55.33 & 31.92  & 46.32     & 49.51 & 41.84  \\
\caa      & 63.96  & 71.80 & 54.15  & 78.16    & 80.92  & 61.61  & 58.42     & 60.70  & 55.01 & 47.97     & 56.01 & 35.10  & 46.17     & 50.92 & 39.52  \\ 
\ours     & \textbf{\begin{tabular}[c]{@{}c@{}}65.22\\ (+3.60)\end{tabular}} & \textbf{\begin{tabular}[c]{@{}c@{}}72.56\\ (+3.56)\end{tabular}} & \textbf{\begin{tabular}[c]{@{}c@{}}56.05\\ (+3.65)\end{tabular}} & \textbf{\begin{tabular}[c]{@{}c@{}}79.92   \\ (+4.96)\end{tabular}} & \textbf{\begin{tabular}[c]{@{}c@{}}82.03\\ (+4.20)\end{tabular}} & \textbf{\begin{tabular}[c]{@{}c@{}}67.24\\ (+9.46)\end{tabular}} & \textbf{\begin{tabular}[c]{@{}c@{}}59.35\rlap{\textsuperscript{$\dagger$}} \\ (+2.30)\end{tabular}} & \textbf{\begin{tabular}[c]{@{}c@{}}62.04\rlap{\textsuperscript{$\dagger$}} \\ (+2.33)\end{tabular}} & \textbf{\begin{tabular}[c]{@{}c@{}}55.32\\ (+2.26)\end{tabular}} & \textbf{\begin{tabular}[c]{@{}c@{}}48.45   \\ (+1.10)\end{tabular} }& \textbf{\begin{tabular}[c]{@{}c@{}}56.45\rlap{\textsuperscript{$\dagger$}}\\ (+1.14)\end{tabular}} & \textbf{\begin{tabular}[c]{@{}c@{}}35.64\\ (+1.02)\end{tabular}} & \textbf{\begin{tabular}[c]{@{}c@{}}48.12   \\ (+1.64)\end{tabular}} & \textbf{\begin{tabular}[c]{@{}c@{}}51.44\\ (+1.40)\end{tabular}} & \textbf{\begin{tabular}[c]{@{}c@{}}43.47\\ (+1.99)\end{tabular}} \\ \midrule
\sft      & 66.89  & 76.84 & 54.45  & 87.36    & 89.50     & 74.52     & 43.78 & 48.63    & 36.50 & 42.18     & 47.95 & 32.96  & 69.68     & 76.76 & 59.76  \\ 
\sft+\ours & \textbf{\begin{tabular}[c]{@{}c@{}}69.24 \\ (+2.35)\end{tabular}} & \textbf{\begin{tabular}[c]{@{}c@{}} 79.28\rlap{\textsuperscript{$\dagger$}}\\ (+2.44)\end{tabular}} & \textbf{\begin{tabular}[c]{@{}c@{}} 61.22 \\ (+6.77)\end{tabular}} & \textbf{\begin{tabular}[c]{@{}c@{}} 88.11\rlap{\textsuperscript{$\dagger$}}\\ (+0.75)\end{tabular}} & \textbf{\begin{tabular}[c]{@{}c@{}}90.00\rlap{\textsuperscript{$\dagger$}}\\ (+0.50)\end{tabular}} & \textbf{\begin{tabular}[c]{@{}c@{}}76.77\\ (+2.25)\end{tabular}} & \textbf{\begin{tabular}[c]{@{}c@{}}49.84 \\ (+6.06)\end{tabular}} & \textbf{\begin{tabular}[c]{@{}c@{}}57.58\\ (+8.95)\end{tabular}} & \textbf{\begin{tabular}[c]{@{}c@{}}38.24 \\ (+1.74)\end{tabular}} & \textbf{\begin{tabular}[c]{@{}c@{}}42.55 \\ (+0.37)\end{tabular} }& \textbf{\begin{tabular}[c]{@{}c@{}}48.38 \\ (+0.43)\end{tabular}} & \textbf{\begin{tabular}[c]{@{}c@{}}33.22 \\ (+0.26)\end{tabular}} & \textbf{\begin{tabular}[c]{@{}c@{}} 71.17\rlap{\textsuperscript{$\dagger$}}\\ (+1.49)\end{tabular}} & \textbf{\begin{tabular}[c]{@{}c@{}}77.83\rlap{\textsuperscript{$\dagger$}}\\ (+1.07)\end{tabular}} & \textbf{\begin{tabular}[c]{@{}c@{}} 61.84\rlap{\textsuperscript{$\dagger$}}\\ (+2.08)\end{tabular}} \\
\bottomrule
\end{tabular}
\caption{\label{tab:main-mc} Main results of discriminative tasks. All the tasks are evaluated using accuracy. \textsuperscript{$\dagger$} denotes the best results. $\avgall$, $\avgseen$, and $\avgunseen$ indicate the macro-average of results across all the languages, the seen languages, and the unseen languages, respectively.
}
\end{table*}

%% file: tables/main-gen.tex
\begin{table*}[t] \scriptsize
\centering
\setlength{\tabcolsep}{3.5pt}
\begin{tabular}{lcccccccccccc}
\toprule
 & \multicolumn{3}{c}{\mzsre}   & \multicolumn{3}{c}{\flores}  & \multicolumn{3}{c}{\wmt}   & \multicolumn{3}{c}{\mgsm}    \\ \cmidrule(rl){2-4} \cmidrule(rl){5-7} \cmidrule(rl){8-10} \cmidrule(rl){11-13}
 &  $\avgall$    & $\avgseen$     & $\avgunseen$ &  $\avgall$    & $\avgseen$     & $\avgunseen$ &  $\avgall$    & $\avgseen$     & $\avgunseen$ &  $\avgall$    & $\avgseen$     & $\avgunseen$ \\\midrule
\baseline & 39.96  & 45.79  & 32.67  & 46.09  & 58.57  & 21.12  & 13.78  & 14.39  & 13.63  & 39.35  & 39.80  & 38.90  \\
\mtgoogle & 73.56\rlap{\textsuperscript{$\dagger$}}   & 72.76\rlap{\textsuperscript{$\dagger$}}   & 74.56\rlap{\textsuperscript{$\dagger$}}   & -   & -   & -   & -   & -   & -   & 46.70\rlap{\textsuperscript{$\dagger$}}  & 47.70\rlap{\textsuperscript{$\dagger$}}  & 45.70\rlap{\textsuperscript{$\dagger$}}  \\
\mtllm    & 33.18  & 39.25  & 25.61  & -   & -   & -   & -   & -   & -   & 21.40  & 30.00  & 12.80  \\
\multicolumn{13}{l}{\cellcolor{gray!15}\textbf{Intervention Methods}}   \\
\iti      & 36.31  & 41.72  & 29.54  & \phantom{0}2.85 & \phantom{0}2.97 & \phantom{0}1.95 & \phantom{0}2.34 &  \phantom{0}3.16 & \phantom{0}2.13    & 40.50  & 41.90  & 39.10  \\
\caa      & 42.88  & 50.17  & 33.78  & 47.87    &  60.63   &    16.75 &  13.74   &  14.86 &  13.46 &  39.43 &  40.85 &  38.00 \\ 
\ours     & \begin{tabular}[c]{@{}c@{}}\textbf{43.22}\\ \textbf{(+3.26)}\end{tabular} & \begin{tabular}[c]{@{}c@{}}\textbf{50.21}\\ \textbf{(+4.42)}\end{tabular} & \begin{tabular}[c]{@{}c@{}}\textbf{34.49}\\ \textbf{(+1.82)}\end{tabular} & 
\begin{tabular}[c]{@{}c@{}}\textbf{48.19}\rlap{\textsuperscript{$\dagger$}}\\ \textbf{(+2.10)}\end{tabular} & \begin{tabular}[c]{@{}c@{}}\textbf{61.28}\rlap{\textsuperscript{$\dagger$}}\\ \textbf{(+2.71)}\end{tabular} & \begin{tabular}[c]{@{}c@{}}\textbf{22.00}\rlap{\textsuperscript{$\dagger$}}\\ \textbf{(+0.88)}\end{tabular} 
& \begin{tabular}[c]{@{}c@{}}\textbf{14.23}\rlap{\textsuperscript{$\dagger$}}\\ \textbf{(+0.45)}\end{tabular} & \begin{tabular}[c]{@{}c@{}}\textbf{15.05}\rlap{\textsuperscript{$\dagger$}}\\ \textbf{(+0.66)}\end{tabular} & \begin{tabular}[c]{@{}c@{}}\textbf{14.02}\rlap{\textsuperscript{$\dagger$}}\\ \textbf{(+0.39)}\end{tabular} & \begin{tabular}[c]{@{}c@{}}\textbf{42.85}\\ \textbf{(+3.50)}\end{tabular} & \begin{tabular}[c]{@{}c@{}}\textbf{43.30}\\ \textbf{(+3.50)}\end{tabular} & \begin{tabular}[c]{@{}c@{}}\textbf{42.40}\\ \textbf{(+3.50)}\end{tabular} \\
\bottomrule
\end{tabular}
\caption{\label{tab:main-gen} Main results of generative tasks. \textsuperscript{$\dagger$} denotes the best results. $\avgall$, $\avgseen$, and $\avgunseen$ indicate the macro-average of results across all the languages, the seen languages, and the unseen languages, respectively. We use Exact Match (EM) to evaluate \mzsre, use BLEU to evaluate \flores and \wmt, and use accuracy to evaluate \mgsm.}
\end{table*}

%% file: 5_analysis_and_discussion.tex
\section{Analysis}
\label{sec:analysis}

In this section, we conduct an in-depth analysis of \ours, focusing on four key aspects: computational costs, enhanced consistency after intervention, the impacts of the intervened components of LLMs, and the choice of intervention strength $\alpha$. This analysis provides a comprehensive understanding of how \ours operates and its implications for model performance and efficiency.

\begin{figure}[t]
\centering          
\subfigure[Training cost]{\label{fig:acc-time}\includegraphics[scale=0.6]{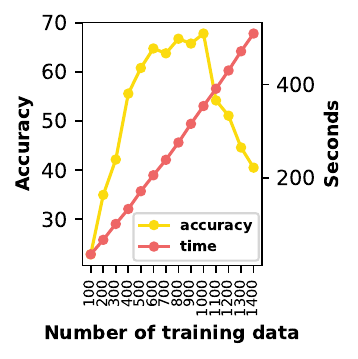}} 
\subfigure[Prediction consistency]{\label{fig:xltr}\includegraphics[scale=0.6]{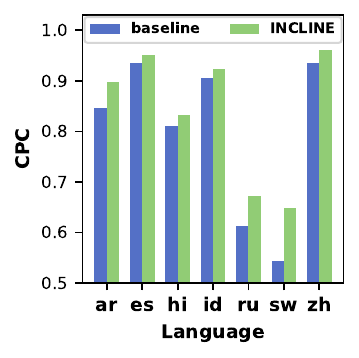}} 
\caption{(a) Training costs of \ours with regard to the number of parallel sentences and time used for training alignment matrices. \ours is evaluated on \xstorycloze in Swahili. (b) Correct Prediction Consistency (CPC) between non-English and English on \xstorycloze for the model using \ours.}
\label{fig:acc-time&xltr}
\end{figure}

\paragraph{\ours is highly efficient for training and introduces only marginal overhead for inference.} 
To analyze the relationship between computational costs and accuracy, we measure both the training and inference costs of our method, \ours, using the \xstorycloze task in Swahili. As shown in \autoref{fig:acc-time}, increasing the amount of training data does not necessarily lead to improved accuracy, even though the training time is directly proportional to the number of samples. In our study, we empirically determine that using 500 samples for training the alignment matrices provides the best balance between performance gains and computational costs. Consequently, the training process takes only 172 seconds. During inference, our approach involves a single intervention at the last token, resulting in a time complexity of $O(1)$. This method incurs only a 12\% increase in inference time, taking 0.80 seconds per item compared to 0.71 seconds without it, thereby maintaining a low inference cost.

\paragraph{\ours effectively enhances the consistency of correct predictions between non-English languages (source) and English (target).} 
Recent non-English test sets are commonly translated from their English versions, either by humans or machines, creating parallel datasets. To quantify the alignment between non-English languages (source) and English (target), we propose using the Correct Prediction Consistency (CPC) rate. This metric measures the proportion of questions correctly answered in both languages, with a higher CPC rate indicating better alignment. The results in \autoref{fig:xltr} demonstrate that CPC significantly improves after intervention by \ours, suggesting that \ours effectively aligns non-English representations with English ones for more accurate predictions. Notably, CPC for Swahili (sw) increases from 0.54 to 0.65 with \ours, showing its effectiveness for low-resource languages.

\input{tables/configuration}

\paragraph{Intervening on hidden states yields the greatest performance improvements.} 
We apply \ours to various components of LLMs, including the hidden states (\ourshidden), the outputs of attention heads (\oursattn), the outputs of FFN blocks (\oursffn), and the embeddings (\oursemb). The results presented in \autoref{tab:config} indicate that intervening on the hidden states (\ourshidden) leads to the most significant improvements across multilingual tasks. This finding suggests that hidden states can capture comprehensive semantic information that is crucial for cross-lingual alignment. While \oursattn, \oursffn, and \oursemb also enhance performance, their performance gains vary across different tasks. These findings justify our design choice of using hidden states in \ours.

\begin{figure}[t]
    \centering
    \includegraphics[scale=0.6]{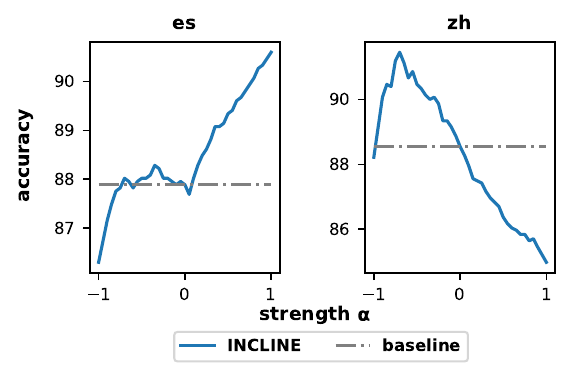}
    \caption{The accuracy changed with hyperparameter $\alpha$ on the \xstorycloze task with \bloomz.}
    \label{fig:parameter}
\end{figure}

\paragraph{The value of $\alpha$ varies across languages and depends on language relatedness.}
In this study, we introduce $\alpha$ to control the strength of intervention in \autoref{eq:mix}. To investigate the impact of $\alpha$, we conduct a grid search to find the optimal $\alpha$ values across the languages in \xstorycloze. We present the results for Spanish and Chinese in \autoref{fig:parameter}. We observe that the optimal $\alpha$ values for these two languages are opposite: positive for Spanish and negative for Chinese. These findings suggest that the value of $\alpha$ is likely to depend on language relatedness, as both Spanish and English belong to the Indo-European language family, while Chinese belongs to the Sino-Tibetan language family. Results for more languages are provided in \autoref{sec:appendix-direction}.

%% file: tables/configuration.tex
\begin{table}[t] \small
\centering
\setlength{\tabcolsep}{4pt}
\begin{tabular}{lcccc}
\toprule
& \xcopa& \xcsqa & \flores & \mgsm  \\ \midrule
\baseline     & 61.60 & 47.35 & 46.09  & 39.35 \\
\multicolumn{5}{l}{\cellcolor{gray!15}\textbf{\ours}}   \\
├ \ourshidden & \textbf{65.22} & \textbf{48.45} & \textbf{48.19}  & \textbf{42.85} \\
├ \oursattn   & 63.87 & 48.18 & 47.54  & 41.55 \\
├ \oursffn & 64.20 & 47.96 & 46.10  & 41.80 \\
└ \oursemb & 63.16 & 47.59 & 39.23 & 40.90 \\
\bottomrule
\end{tabular}
\caption{\label{tab:config} The averaged results of \xstorycloze, \xcsqa, \flores, \mgsm  tasks with four configurations for \ours given by \bloomz.}
\end{table}

%% file: 6_discussion.tex
\input{tables/llms}
\input{tables/ood}
\section{Discussions}
\label{sec:discussions}
In this section, we conduct a series of experiments to investigate how variations in LLMs, model sizes, in-context learning, and the data used for training alignment matrices affect our results. Additionally, we also explore using French as the target language (\autoref{sec:appendix-fr}) and examine the effects of layer-specific intervention (\autoref{sec:appendix-layerwise}).

\paragraph{\ours consistently enhances performance across multiple LLMs.}
To demonstrate the versatility of \ours across different LLMs, we apply it to another four high-performing models on the \xstorycloze task. As shown in \autoref{tab:otherllms}, \ours consistently enhances performance compared to the \baseline. Specifically, we observe increases of +4.96 for \bloomz, +5.40 for \llamathree, +7.03 for \llama, +8.64 for \mistral, and +2.74 for \falcon.

\begin{figure}[t]
\centering          
\subfigure[Various model sizes]{\label{fig:scale}\includegraphics[scale=0.51]{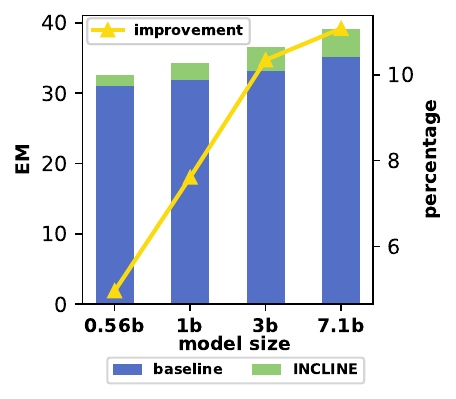}} 
\subfigure[In-context learning]{\label{fig:fewshots}\includegraphics[scale=0.59]{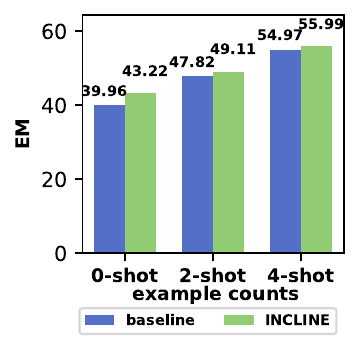}} 
\caption{(a) Exact Match (left y-axis) and relative improvements over the baseline (right y-axis) on \mzsre with respect to various model sizes of \textsc{BLOOMZ}. (b) Exact Match score for \mzsre dataset with \ours based on the zero-shot setting and few-shot settings given by \bloomz.}
\label{fig:scale_fewshots}
\end{figure}

\paragraph{Larger LLMs benefit more from \ours.} 
Building on the work of \citet{mzsre}, who demonstrates a scaling relationship between the size of backbone models and their performance, we evaluate the impact of different model sizes within the \textsc{BLOOMZ} series on the \mzsre dataset. Our findings, illustrated in \autoref{fig:scale}, show that the relative performance gain of \ours over the baseline increases with the size of the backbone model. Specifically, the Exact Match (EM) scores (in the stacked columns) and the improvement percentages (in the line chart) indicate that larger models in the \textsc{BLOOMZ} series exhibit more significant enhancements when \ours is applied. This observation demonstrates that larger LLMs can benefit more from \ours.

\paragraph{\ours can further enhance model performance when combined with in-context learning.}
In-context learning (ICL) has been shown to improve the performance of LLMs on the \mzsre task \citep{mzsre}. Building upon this finding, we evaluate the effectiveness of combining \ours with ICL. As illustrated in \autoref{fig:fewshots}, \ours demonstrates enhanced performance, achieving an additional increase of +1.02 in average Exact Match (EM) score with four in-context examples compared to the baseline using ICL alone. While this improvement is smaller than the +3.26 increase observed in the zero-shot setting, it suggests that the benefits of \ours and ICL are complementary, with both methods capturing features from different perspectives. This highlights the versatility of \ours in various applications.

\paragraph{High-quality parallel sentences improve alignment in \ours.}
We explore how the quality of parallel sentences affects the performance of \ours. By default, the alignment matrices of \ours are trained using 500 random samples from the News Commentary dataset. To assess the impact of sentence quality, we also train the alignment matrices using 500 high-quality parallel sentences from the development set of \flores, which are carefully translated by professional human translators. We refer to this variant as \oursfdev. In \autoref{tab:quality}, \oursfdev significantly outperforms both the standard \ours and \baseline, highlighting the importance of high-quality parallel sentences.

%% file: tables/llms.tex
\begin{table}[t] \scriptsize
\centering
\setlength{\tabcolsep}{3pt}
\begin{tabular}{lcccccccc}
\toprule
& \textbf{ar}    & \textbf{es}    & \textbf{hi}    & \textbf{id}    & \textbf{ru}   & \textbf{sw}      & \textbf{zh}    & \textbf{AVG}   \\ \midrule
\multicolumn{9}{c}{\cellcolor{gray!15}\textbf{\bloomz}}   \\
\textbf{\baseline} & 79.22 & 87.89 & 76.37 & 84.45 & 57.78 & 50.50 & 88.55 & 74.96 \\
\textbf{\ours} & \textbf{83.12} & \textbf{90.60} & \textbf{81.47} &\textbf{ 86.10} & \textbf{67.24} & \textbf{59.70} & \textbf{91.20} & \textbf{79.92} \\
\multicolumn{9}{c}{\cellcolor{gray!15}\textbf{\llamathree}}   \\
\textbf{\baseline} & 86.50 & 91.73 & 84.84 & 37.46 & 66.98 & 54.00 & 92.39 & 73.41 \\
\textbf{\ours} & \textbf{87.36} & \textbf{92.39} & \textbf{85.31} & \textbf{64.53} & \textbf{73.73} & \textbf{55.66} & \textbf{92.72} & \textbf{78.81} \\ 
\multicolumn{9}{c}{\cellcolor{gray!15}\textbf{\llama}}   \\
\textbf{\baseline} & 49.37 & 47.25 & 39.25 & 48.18 & 34.94 & \phantom{0}0.93  & 55.53 & 39.35 \\
\textbf{\ours} & \textbf{51.42} & \textbf{56.65} & \textbf{47.25} & \textbf{49.97} & \textbf{41.03} & \textbf{17.67} & \textbf{60.69} & \textbf{46.38} \\
\multicolumn{9}{c}{\cellcolor{gray!15}\textbf{\mistral}}   \\
\textbf{\baseline} & 18.33 & 81.34 & 24.95 & 76.64 & 83.65 & \phantom{0}2.58  & 90.07 & 53.94 \\
\textbf{\ours} & \textbf{36.71} & \textbf{84.23} & \textbf{35.77} & \textbf{80.18} & \textbf{85.13} & \textbf{25.71} & \textbf{90.34} & \textbf{62.58} \\ 
\multicolumn{9}{c}{\cellcolor{gray!15}\textbf{\falcon}}   \\
\textbf{\baseline} & 53.61 & 58.31 & 53.21 & 55.59 & 54.60 & 51.16 & 54.00 & 54.35 \\
\textbf{\ours} & \textbf{54.33} & \textbf{61.81} & \textbf{54.33} & \textbf{58.04} & \textbf{57.91} & \textbf{53.47} & \textbf{59.70} & \textbf{57.09} \\ 
\bottomrule
\end{tabular}
\caption{\label{tab:otherllms} The results of \xstorycloze dataset with five LLM backbones.}
\end{table}

%% file: tables/ood.tex
\begin{table*}[t]\small
\centering
\begin{tabular}{lcccccccccc}
\toprule
\textbf{} & \textbf{ar}   & \textbf{el}   & \textbf{es}   & \textbf{fr}   & \textbf{hi}   & \textbf{ru}   & \textbf{tr}   & \textbf{vi}   & \textbf{zh}   & \textbf{AVG}  \\ \midrule

\textbf{\baseline} & 66.59 & 15.30 & 48.52 & 67.86 & 71.97 & 35.66 & 12.38 & 40.40 & 56.11 & 46.09 \\
\textbf{\ours}   & 68.68 & 15.63 & 50.79 & 69.93 & 76.92 & 37.95 & 12.42 & 43.11 & 58.27 & 48.19 \\
\textbf{\oursfdev}   & \textbf{73.95} &\textbf{ 15.76} &\textbf{ 56.11} & \textbf{75.84} & \textbf{77.85 }& \textbf{39.33} & \textbf{12.92 }&\textbf{ 46.49} & \textbf{60.19} & \textbf{50.94} \\
\bottomrule
\end{tabular}
\caption{\label{tab:quality} The BLEU results of \flores dataset with \ours and \oursfdev.}
\end{table*}

%% file: 7_conclusion.tex
\section{Conclusion}
\label{sec:conclusion}

In this paper, we introduce \textbf{\oursfull (\ours)}, an innovative framework that bridges the performance gaps between high-performing and low-performing languages in LLMs. By training alignment matrices to transform source low-performing language representations into the target high-performing language representation space, \ours enhances performance on underrepresented languages without requiring additional training or fine-tuning of LLMs. Extensive experiments across nine benchmarks and five LLMs demonstrate that, \ours delivers significant improvements by up to +4.96 in terms of accuracy compared to strong baselines, while it only requires minimal computational costs.

%% file: 9_limitations.tex
\section{Limitations}

While \ours demonstrates significant enhancement for the multilingual tasks with cross-lingual intervention, the alignment matrices are trained for specific pairs of source and target languages. 
Future work will focus on developing multilingual alignment matrices that can accommodate multiple languages simultaneously, reducing the need for language pair-specific training and enhancing scalability.
Implementing \ours requires access to the internal layers and representations of LLMs. For proprietary or closed-source models, or models accessible only through APIs without exposure of internal representations (e.g., GPT-4o), applying this method may not be feasible. How to perform cross-lingual alignment as a plug-and-play tool for all LLMs, including those with restricted access, requires further investigation. 

%% file: 8_appendix.tex
\section{Details of Datasets}
\label{sec:appendix-dataset}
The tasks and the corresponding output format, prompt template, evaluation metrics, the number of languages are shown in \autoref{tab:dataset}.

\input{tables/data}

\section{Hyperparameters for \sft}
\label{sec:appendix-sft_hyperparameters}
We fine-tune all parameters of LLMs using the AdamW optimizer with a learning rate of $2 \times 10^{-6}$ and a batch size of 4. This process is conducted over three epochs on four NVIDIA A100 GPUs (80GB). During training, we use a linear learning rate schedule with a warm-up phase that constitutes 10\% of the total training steps.

\section{Detailed Results of Intervention}
\label{sec:appendix-detail}

\input{tables/main}
The detailed results of \baseline, \mtgoogle, \mtllm, \sft, \iti, \caa \ours and \sft+\ours for each languages across discriminative and generative tasks are shown in \autoref{tab:overall}.

\section{The value of $\alpha$ across languages}
\label{sec:appendix-direction}
\begin{figure*}[h]
    \centering
    \includegraphics[scale=0.65]{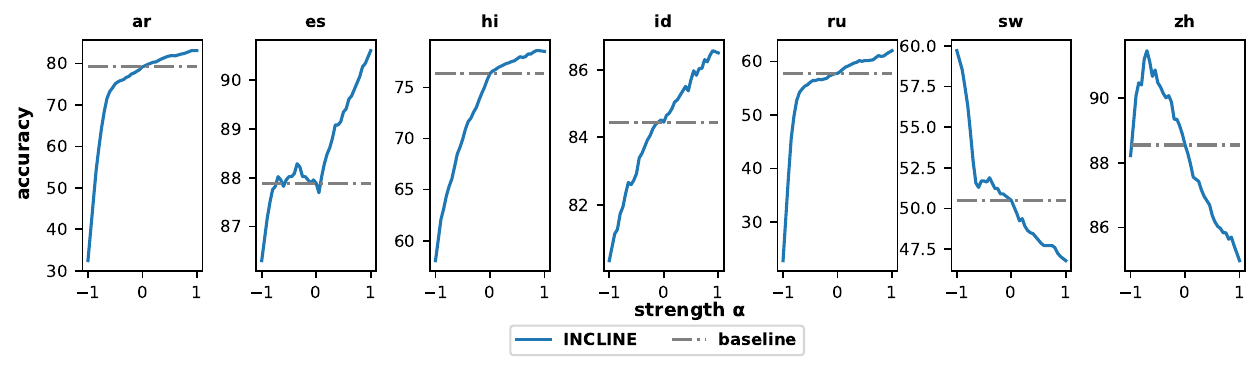}
    \caption{The accuracy changed with hyperparameter $\alpha$ on the \xstorycloze task.}
    \label{fig:parameter-alllangs}
\end{figure*}

We explore the optimal value of $\alpha$ for each language in \xstorycloze using grid search, as shown in \autoref{fig:parameter-alllangs}.

\section{Projection to Non-English}
\label{sec:appendix-fr}
\input{tables/fr}

We have demonstrated the effectiveness of \ours in aligning representations from non-English to English. To further prove the generalizability of \ours with another high-performing language, we conduct an ablation study aligning representations of various languages with French. As shown in \autoref{tab:xx2fr}, \ours enhances translation performance to non-English languages, with an average BLEU score increase of +5.35. This further demonstrates that \ours can effectively align representations across different languages.

\section{Layer-Specific Intervention}
\label{sec:appendix-layerwise}
\begin{figure}[h]
    \centering
    \includegraphics[scale=0.65]{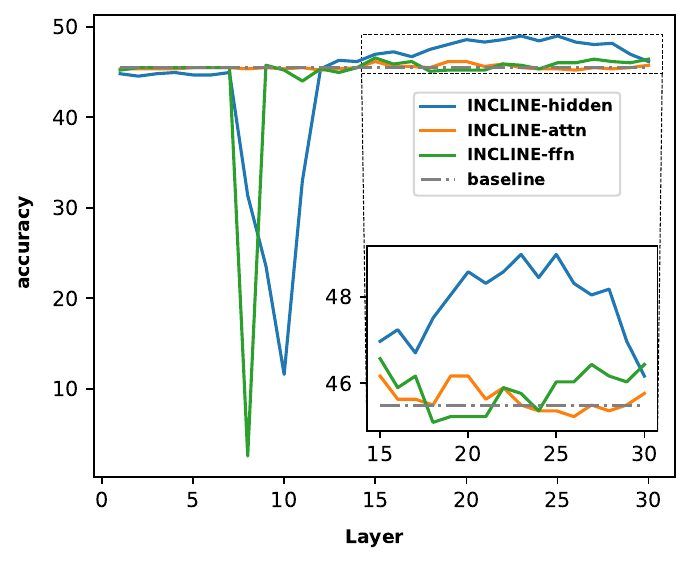}
    \caption{The accuracy changed with layer-specific intervention, where \ours-hidden, \ours-attn, \ours-ffn represents the intervention in the hidden states, in the output of attention heads, in the output of the FFN block.}
    \label{fig:layer-wise}
\end{figure}

To examine the effects of layer-specific interventions, we conduct a study applying interventions across different layers and evaluated the results using the \mzsre Portuguese test set. The findings, shown in \autoref{fig:layer-wise}, demonstrate how accuracy varies with interventions at different layers. Intervening in a single layer (scoring less than 50) resulted in lower performance compared to interventions across all layers (52.09 in \autoref{tab:overall}). According to the trends in \autoref{fig:layer-wise}, interventions in the higher layers lead to greater improvements than those in the lower layers, likely because they mitigate information forgetting. Notably, interventions in the hidden states outperform other types significantly. However, not every intervention leads to performance gains; both \ourshidden and \oursffn show substantial declines when intervening in the middle layers. The mechanisms underlying these effects merit further investigation.

\section{Details of Visualizing}
\label{sec:appendix-visualize}

Following \citet{iti}, we use Linear Regression to examine multilingual input representations. For each English and corresponding Portuguese sample from the News Commentary dataset (a total of 500 items), we extract the hidden states at the last token to create a probing dataset for each layer. We randomly divide this dataset into training and validation sets in a 4:1 ratio and fit a binary linear classifier to the training set. Similar to principal component analysis (PCA), we train a second linear probe on the same dataset, constrained to be orthogonal to the first probe. This orthogonality ensures that the two probes capture distinct aspects of the data. Finally, we project the hidden states of each sample in the \mzsre test set onto the directions defined by the probes from the last layer, allowing us to visualize and analyze the multilingual representations effectively.

%% file: tables/data.tex
\begin{table*}[h]\small
\centering
\begin{tabular}{lllll}
\toprule
Dataset     & Output & prompt & Metric & |L| \\ \midrule
\xcopa & 2-way class & \begin{tabular}[c]{@{}l@{}}Here is a premise: "\{premise\}". A: "\{choice1\}" B: "\{choice2\}" \\ What is the \{question\}? "A" or "B"?\end{tabular}  & acc.  & 10 \\  \midrule
\xstorycloze & 2-way class  & \begin{tabular}[c]{@{}l@{}}\{input\} What is a possible continuation for the story given the \\ following options? A: \{quiz1\} B: \{quiz2\}'\end{tabular}& acc.  & 8 \\ \midrule
\xwinograd   & 2-way class & \begin{tabular}[c]{@{}l@{}}\{input\} Replace the \_ in the above sentence with the correct option:\\ - \{option1\} - \{option2\}\end{tabular}   & acc.  & 6 \\ \midrule
\xnli  & 3-way class  & \begin{tabular}[c]{@{}l@{}}Take the following as truth: \{premise\} Then the following statement: \\ "\{hypothesis\}" is "true", "false", or "inconclusive"?\end{tabular} & acc.  & 13 \\ \midrule
\xcsqa & multi-choice & Question: \{question\} \{choice\} Answer:  & acc.   
 & 14 \\ \midrule
\mzsre & answer & \{context\} Quesion: \{question\} Answer: & EM  &  10  \\ \midrule
\flores & answer & Translate the following sentence from \{language\} to English: \{input\} & BLEU  & 10 \\ \midrule
\wmt & answer & Translate the following sentence from \{language\} to English: \{input\} & BLEU  & 5 \\ \midrule
\mgsm & answer & \begin{tabular}[c]{@{}l@{}}Write a response that appropriately completes the request in \{language\}.\\Please answer in \{language\}. \#\#\# Instruction: \{query\}\#\#\# Response:\end{tabular}& EM & 9 \\
\bottomrule
\end{tabular}
\caption{\label{tab:dataset} The nine datasets used to evaluate multilingual intervention. |L| indicates the number of languages. EM is the Exact Match score and acc. represents the accuracy.}
\end{table*}

%% file: tables/main.tex
\begin{table*}[htbp] \scriptsize
\centering
\setlength{\tabcolsep}{4pt}
\begin{tabular}{lccccccccccccccc}
\toprule
\multicolumn{16}{c}{\textbf{Discriminative tasks}}  \\ \midrule
\rowcolor[gray]{.93} \textbf{\xcopa}      & \textbf{en} & \textbf{et}    & \textbf{id}    & \textbf{it}    & \textbf{sw}    & \textbf{ta}    & \textbf{th}    & \textbf{tr}    & \textbf{vi}    & \textbf{zh}    & \textbf{}      & \textbf{}      & \textbf{}      & \textbf{}      & \textbf{AVG}   \\
\textbf{\baseline}    & 76.40       & 50.80   & 69.60   & 58.60   & 55.20   & 71.60   & 50.60   & 49.60   & 71.20   & 77.40   &     &     &     &     & 61.62   \\
\textbf{\mtgoogle}   & - & 75.40\rlap{\textsuperscript{$\dagger$}}  & 75.00   & 76.00\rlap{\textsuperscript{$\dagger$}}   & 76.20\rlap{\textsuperscript{$\dagger$}}   & 62.20   & 62.40\rlap{\textsuperscript{$\dagger$}}   & 78.40\rlap{\textsuperscript{$\dagger$}}   & 76.40   & 77.80   &     &     &     &     & 73.31\rlap{\textsuperscript{$\dagger$}}   \\
\textbf{\mtllm}     & - & 44.80   & 69.80   & 59.40   & 60.20   & 71.20   & 47.40   & 51.20   & 61.60   & 73.00   &     &     &     &     & 59.84   \\
\textbf{\sft}        & 86.40       & 50.60   & 78.40  & 67.80   & 59.00   & 77.20   & 47.60   & 53.00   & 83.00   & 84.60   &     &     &     &     & 66.80   \\
\textbf{\iti}  & - & 50.80   & 70.80   & 60.00   & 55.40   & 63.20   & 49.00   & 50.60   & 69.00   & 79.40   &     &     &     &     & 60.91   \\ 
\textbf{\caa}    & - &  51.20 & 	72.20 & 	61.20 & 	59.20 & 	73.00 & 	52.20 & 	52.00 & 	74.80 & 	79.80 &  &   &   &   & 63.96  \\ 
\cdashline{1-16}[1pt/1pt]
\textbf{\ours} & - & \textbf{55.40} & \textbf{73.40} & \textbf{62.80} & \textbf{59.80} & \textbf{73.40} & \textbf{52.60} & \textbf{53.40} & \textbf{76.20} & \textbf{80.00} & \textbf{}      & \textbf{}      & \textbf{}      & \textbf{}      & \textbf{65.22}\\ 
\textbf{\sft+\ours} & - & \textbf{53.20} & \textbf{81.20}\rlap{\textsuperscript{$\dagger$}} & \textbf{65.80} & \textbf{60.80} & \textbf{85.00}\rlap{\textsuperscript{$\dagger$}} & \textbf{54.40 }& \textbf{53.40} &\textbf{84.40}\rlap{\textsuperscript{$\dagger$}} & \textbf{85.00}\rlap{\textsuperscript{$\dagger$}} & \textbf{}      & \textbf{}      & \textbf{}      & \textbf{}      & \textbf{69.24} \\ \hline
\rowcolor[gray]{.93} \textbf{\xstorycloze}  & \textbf{en} & \textbf{ar}    & \textbf{es}    & \textbf{hi}    & \textbf{id}    & \textbf{ru}    & \textbf{sw}    & \textbf{zh}    & \textbf{}      & \textbf{}      & \textbf{}      & \textbf{}      & \textbf{}      & \textbf{}      & \textbf{AVG}   \\
\textbf{\baseline}    & 91.46       & 79.22   & 87.89   & 76.37   & 84.45   & 57.78   & 50.50   & 88.55   &     &     &     &     &     &     & 74.96   \\
\textbf{\mtgoogle}   & - & 79.48   & 81.34   & 50.69   & 80.81   & 80.08\rlap{\textsuperscript{$\dagger$}}   & 77.04   & 86.96   &     &     &     &     &     &     & 76.63   \\
\textbf{\mtllm}     & - & 81.80   & 86.83   & 82.59   & 83.59   & 62.48   & 73.66   & 84.91   &     &     &     &     &     &     & 79.41   \\
\textbf{\sft}        & 94.11      & 90.47   & 92.85\   & 88.22   & 91.59   & 74.52   & 81.14   & 92.72   &     &     &     &     &     &     & 87.36   \\
\textbf{\iti}         & - & 78.23   & 90.54   & 80.28   & 85.70   & 58.70   & 52.55   & 88.68   &     &     &     &     &     &     & 76.38   \\ 
\textbf{\caa}    & - &  86.04 & 	90.47 & 	79.15 & 	88.22 & 	61.61 & 	52.61 & 	89.01 &     &     &     &     &     &    	 & 78.16  \\ 
\cdashline{1-16}[1pt/1pt]
\textbf{\ours} & - & \textbf{83.12} & \textbf{90.60} & \textbf{81.47} &\textbf{ 86.10} & \textbf{67.24} & \textbf{59.70} & \textbf{91.20} & \textbf{}      & \textbf{}      & \textbf{}      & \textbf{}      & \textbf{}      & \textbf{}      & \textbf{79.92} \\ 
\textbf{\sft+\ours} & -  & \textbf{90.93}\rlap{\textsuperscript{$\dagger$}} & \textbf{92.98}\rlap{\textsuperscript{$\dagger$}} & \textbf{89.08}\rlap{\textsuperscript{$\dagger$}} & \textbf{91.99}\rlap{\textsuperscript{$\dagger$}} & \textbf{76.77} & \textbf{81.93}\rlap{\textsuperscript{$\dagger$}} & \textbf{93.05}\rlap{\textsuperscript{$\dagger$}} & \textbf{}      & \textbf{}      & \textbf{}      & \textbf{}      & \textbf{}      & \textbf{}      &  \textbf{88.11}\rlap{\textsuperscript{$\dagger$}} \\ \hline
\rowcolor[gray]{.93} \textbf{\xwinograd}    & \textbf{en} & \textbf{fr}    & \textbf{ja}    & \textbf{pt}    & \textbf{ru}    & \textbf{zh}    & \textbf{}      & \textbf{}      & \textbf{}      & \textbf{}      & \textbf{}      & \textbf{}      & \textbf{}      & \textbf{}      & \textbf{AVG}   \\
\textbf{\baseline}    & 73.76       & 59.04   & 51.51   & 57.80   & 54.60   & 62.30   &     &     &     &     &     &     &     &     & 57.05   \\
\textbf{\mtgoogle}   & - & 61.45   & 58.39\rlap{\textsuperscript{$\dagger$}}   & 59.32\rlap{\textsuperscript{$\dagger$}}   & 57.41  & 50.60   &     &     &     &     &     &     &     &     & 57.63   \\
\textbf{\mtllm}     & - & 54.22   & 47.86   & 33.08   & 42.22   & 37.70   &     &     &     &     &     &     &     &     & 43.02   \\
\textbf{\sft}  &   78.06   & 62.65	& 14.91	& 43.35	& 58.09	& 39.89	  &     &     &     &     &     &     &     &     & 43.78   \\
\textbf{\iti}         & - & 54.22   & 51.51   & 57.79   & 14.60   & 63.10   &     &     &     &     &     &     &     &     & 48.24   \\ 
\textbf{\caa}    & - & 60.24 &   	52.87 &   58.17 &  57.14 & 	63.69 &    &     &     &     &     &     &     &     &   58.42  \\ 
\cdashline{1-16}[1pt/1pt]
\textbf{\ours} & - & \textbf{63.86}\rlap{\textsuperscript{$\dagger$}} & \textbf{53.18} & \textbf{58.56} & \textbf{57.46} & \textbf{63.69}\rlap{\textsuperscript{$\dagger$}} & \textbf{}      & \textbf{}      & \textbf{}      & \textbf{}      & \textbf{}      & \textbf{}      & \textbf{}      & \textbf{}      & \textbf{59.35}\rlap{\textsuperscript{$\dagger$}}  \\ 
\textbf{\sft+\ours} & - & \textbf{63.86}\rlap{\textsuperscript{$\dagger$}} & \textbf{16.48} & \textbf{46.39} & \textbf{60.00}\rlap{\textsuperscript{$\dagger$}} & \textbf{62.50} & \textbf{}      & \textbf{}      & \textbf{}      & \textbf{}      & \textbf{}      & \textbf{}      & \textbf{}      & \textbf{}      & 	\textbf{49.84}   \\ \hline
\rowcolor[gray]{.93} \textbf{\xcsqa}        & \textbf{en} & \textbf{ar}    & \textbf{de}    & \textbf{es}    & \textbf{fr}    & \textbf{hi}    & \textbf{it}    & \textbf{ja}    & \textbf{nl}    & \textbf{pt}    & \textbf{ru}    & \textbf{sw}    & \textbf{vi}    & \textbf{zh}    & \textbf{AVG}   \\ 
\textbf{\baseline}    & 76.50       & 52.40   & 33.90   & 64.30   & 63.30   & 48.50   & 41.30   & 36.00   & 28.70   & 61.30   & 33.20   & 40.50   & 55.20   & 57.00   & 47.35   \\
\textbf{\mtgoogle}   & - & 61.60\rlap{\textsuperscript{$\dagger$}}   & 65.00\rlap{\textsuperscript{$\dagger$}}   & 68.00\rlap{\textsuperscript{$\dagger$}}  & 67.20\rlap{\textsuperscript{$\dagger$}}   & 32.10   & 68.70\rlap{\textsuperscript{$\dagger$}}   & 57.30\rlap{\textsuperscript{$\dagger$}}   & 66.50\rlap{\textsuperscript{$\dagger$}}   & 66.90\rlap{\textsuperscript{$\dagger$}}   & 64.50\rlap{\textsuperscript{$\dagger$}}   & 19.60   & 62.90\rlap{\textsuperscript{$\dagger$}}   & 60.40\rlap{\textsuperscript{$\dagger$}}   & 58.52\rlap{\textsuperscript{$\dagger$}}   \\
\textbf{\mtllm}     & - & 32.30   & 26.30   & 42.70   & 42.30   & 30.40   & 25.60   & 25.60   & 17.40   & 39.90   & 21.60   & 24.00   & 31.60   & 39.80   & 30.73   \\
\textbf{\sft}        & 65.70       & 48.20   & 32.90   & 54.10   & 53.60   & 43.10   & 40.40   & 32.60   & 29.00   & 53.60   & 29.90   & 31.80   & 48.40   & 50.80   & 42.18   \\
\textbf{\iti}         &  - & 52.10   & 34.20   & 64.50   & 63.70   & 48.10   & 40.00   & 25.90   & 26.00   & 61.20   & 33.50   & 40.90   & 54.90   & 57.20   & 46.32   \\ 
\textbf{\caa}    & - & 52.80 & 	34.10 & 	64.50 & 	63.30 & 	48.40 & 	42.20 & 	36.40 & 	29.30 & 	62.80 & 	33.50 & 	41.90 & 	56.00 & 	58.40 & 	47.97 \\ 
\cdashline{1-16}[1pt/1pt]
\textbf{\ours} & - & \textbf{53.20} & \textbf{34.90} & \textbf{65.00} & \textbf{63.80} & \textbf{48.80}\rlap{\textsuperscript{$\dagger$}} & \textbf{42.90} & \textbf{36.80} & \textbf{29.80} & \textbf{62.60} & \textbf{33.80} & \textbf{42.20}\rlap{\textsuperscript{$\dagger$}} & \textbf{57.30} & \textbf{58.70} & \textbf{48.45} \\ 
\textbf{\sft+\ours} & - & \textbf{48.50} & \textbf{33.30} & \textbf{54.40} & \textbf{53.70} & \textbf{43.90} & \textbf{40.60} & \textbf{33.00} & \textbf{29.30} & 5\textbf{3.70} & \textbf{29.90} & \textbf{32.50} & \textbf{49.10} & \textbf{51.20} & \textbf{42.55}  \\ \hline
\rowcolor[gray]{.93} \textbf{\xnli}         & \textbf{en} & \textbf{ar}    & \textbf{de}    & \textbf{el}    & \textbf{es}    & \textbf{fr}    & \textbf{hi}    & \textbf{ru}    & \textbf{sw}    & \textbf{th}    & \textbf{tr}    & \textbf{vi}    & \textbf{zh}    & \textbf{}      & \textbf{AVG}   \\
\textbf{\baseline}    & 54.81       & 53.63   & 43.33   & 41.04   & 51.36   & 50.54   & 50.16   & 47.80   & 45.01   & 40.32   & 34.93   & 49.68   & 49.92   &     & 46.48   \\
\textbf{\mtgoogle}   & - & 51.46   & 53.13   & 52.71   & 51.84   & 50.82   & 41.58   & 51.68  & 50.54   & 50.50   & 52.00   & 51.94   & 50.42   &     & 50.72  \\
\textbf{\mtllm}     & - & 46.87   & 43.25   & 36.29   & 52.12   & 51.40   & 45.31   & 42.08   & 43.43   & 34.07   & 33.17   & 47.23   & 48.42   &     & 43.64   \\
\textbf{\sft}        & 86.37 & 	77.17 & 	68.10  & 	59.48	 & 82.71	 & 81.48	 & 72.42	 & 66.87 & 67.15	 & 54.55	 & 49.80 & 77.62 & 78.76	 &  & 69.68 \\
\textbf{\iti}         & - & 53.69   & 45.37   & 41.36   & 50.18   & 51.20   & 50.34   & 47.74   & 43.35   & 38.98   & 35.77   & 48.96   & 48.86   &     & 46.32   \\ 
\textbf{\caa}    & - & 53.59	 & 44.67 & 	41.62	 & 52.83 & 	52.75 & 	50.28 & 	34.40 & 	45.75 & 	40.48 & 	36.41 & 	50.32 & 	50.92 &  & 	46.17\\ 
\cdashline{1-16}[1pt/1pt]
\textbf{\ours} & - & \textbf{53.89} & \textbf{47.74} & \textbf{41.96} & \textbf{54.33} &\textbf{ 53.11} & \textbf{50.50}\rlap{\textsuperscript{$\dagger$}} & \textbf{49.22} & \textbf{45.99} & \textbf{41.28} &\textbf{ 37.17} & \textbf{51.12} & \textbf{51.16}\rlap{\textsuperscript{$\dagger$}} & \textbf{}      & \textbf{48.12} \\ 
\textbf{\sft+\ours} & - & \textbf{78.44}\rlap{\textsuperscript{$\dagger$}} & \textbf{71.02}\rlap{\textsuperscript{$\dagger$}} & \textbf{61.22}\rlap{\textsuperscript{$\dagger$}} & \textbf{83.07}\rlap{\textsuperscript{$\dagger$}} & 8\textbf{2.14}\rlap{\textsuperscript{$\dagger$}} & \textbf{73.85}\rlap{\textsuperscript{$\dagger$}} & \textbf{69.68}\rlap{\textsuperscript{$\dagger$}} & \textbf{69.14}\rlap{\textsuperscript{$\dagger$}} & 5\textbf{5.69}\rlap{\textsuperscript{$\dagger$}} & \textbf{51.60}\rlap{\textsuperscript{$\dagger$}} & \textbf{78.64}\rlap{\textsuperscript{$\dagger$}} & \textbf{79.52}\rlap{\textsuperscript{$\dagger$}} &  & \textbf{71.17}\rlap{\textsuperscript{$\dagger$}} \\ \midrule
\multicolumn{16}{c}{\textbf{Generative tasks}}  \\ \midrule
\rowcolor[gray]{.93} \textbf{\mzsre}        & \textbf{en} & \textbf{de}    & \textbf{es}    & \textbf{fr}    & \textbf{pt}    & \textbf{ru}    & \textbf{th}    & \textbf{tr}    & \textbf{vi}    & \textbf{zh}    & \textbf{}      & \textbf{}      & \textbf{}      & \textbf{}      & \textbf{AVG}   \\
\textbf{\baseline}    & 96.23       & 55.05   & 48.86   & 49.53   & 45.49   & 30.55   & \phantom{0}6.33& 38.76   & 51.68   & 33.38   &     &     &     &     & 39.96   \\
\textbf{\mtgoogle}  & - & 78.73\rlap{\textsuperscript{$\dagger$}} & 	76.18\rlap{\textsuperscript{$\dagger$}} & 	75.50\rlap{\textsuperscript{$\dagger$}} & 	71.74\rlap{\textsuperscript{$\dagger$}} & 	63.66\rlap{\textsuperscript{$\dagger$}} & 	78.47\rlap{\textsuperscript{$\dagger$}} & 	77.39\rlap{\textsuperscript{$\dagger$}} & 	60.97\rlap{\textsuperscript{$\dagger$}} & 	79.41\rlap{\textsuperscript{$\dagger$}} &    &     &     &   	& 73.56\rlap{\textsuperscript{$\dagger$}}  \\
\textbf{\mtllm}    & - & 49.13& 	54.78& 	51.28& 	\phantom{0}6.86& 	\phantom{0}2.69& 	\phantom{0}9.69& 	40.92& 	34.72& 	48.59&    &     &     &   	& 33.18 \\
\textbf{\iti}         & - & 53.84   & 44.41   & 43.34   & 41.99   & 19.11   & \phantom{0}6.59& 38.63   & 46.70   & 32.17   &     &     &     &     & 36.31   \\ 
\textbf{\caa}    & - &  57.07 & 	53.30 & 	52.36 & 	52.76 & 	31.49 & 	\phantom{0}7.13 & 	39.43 & 	55.05 & 	38.36 &    &     &     &   	 & 42.99  \\ 
\cdashline{1-16}[1pt/1pt]
\textbf{\ours} & - & \textbf{57.20} & \textbf{53.30} & \textbf{51.82} & \textbf{52.09} & \textbf{31.49} & \phantom{0}\textbf{7.40} & \textbf{41.86} & \textbf{55.32} & \textbf{38.49} & \textbf{}      & \textbf{}      & \textbf{}      & \textbf{}      & \textbf{43.22} \\ \hline
\rowcolor[gray]{.93} \textbf{\flores}       & \textbf{en} & \textbf{ar}    & \textbf{el}    & \textbf{es}    & \textbf{fr}    & \textbf{hi}    & \textbf{ru}    & \textbf{tr}    & \textbf{vi}    & \textbf{zh}    & \textbf{}      & \textbf{}      & \textbf{}      & \textbf{}      & \textbf{AVG}   \\
\textbf{\baseline}    & -       & 66.59& 15.30& 48.52& 67.86& 71.97& 35.66& 12.38& 40.40& 56.11
&     &     &     &     & 46.09
\\
\textbf{\iti}         & - & \phantom{0}2.39& \phantom{0}2.34& \phantom{0}3.71& \phantom{0}4.40& \phantom{0}3.31& \phantom{0}2.44& \phantom{0}3.03& \phantom{0}3.64& \phantom{0}0.37&     &     &     &     & 2.85\\ 
\textbf{\caa}    & - & 67.88 & 15.92 & 54.85 & 68.16 & 72.98 & 38.99 & 12.09 & 43.01 & 56.93 &     &     &     &     & 47.87 \\ 
\cdashline{1-16}[1pt/1pt]
\textbf{\ours} & - & \textbf{73.95}\rlap{\textsuperscript{$\dagger$}}& \textbf{15.79}\rlap{\textsuperscript{$\dagger$}}& \textbf{56.11}\rlap{\textsuperscript{$\dagger$}}& \textbf{75.84}\rlap{\textsuperscript{$\dagger$}}& \textbf{77.85}\rlap{\textsuperscript{$\dagger$}}& \textbf{39.33}\rlap{\textsuperscript{$\dagger$}}& \textbf{12.92}\rlap{\textsuperscript{$\dagger$}}& \textbf{48.62}\rlap{\textsuperscript{$\dagger$}}& \textbf{60.19}\rlap{\textsuperscript{$\dagger$}}
& \textbf{}      & \textbf{}      & \textbf{}      & \textbf{}      & \textbf{51.18}\rlap{\textsuperscript{$\dagger$}}
\\ \hline
\rowcolor[gray]{.93} \textbf{\wmt}       & \textbf{en} & \textbf{de}    & \textbf{ja}    & \textbf{ru}    & \textbf{uk}    & \textbf{zh}    &    &    &    &    &      &     &     &     & \textbf{AVG}   \\
\textbf{\baseline}    & -       & 18.26	  & 10.17  & 14.73  & 11.36  & 14.39   &    &    &    &    &      &     &     &     & 11.78 \\
\textbf{\iti}         & - &  \phantom{0}2.75 & \phantom{0}1.79 & 	\phantom{0}2.32 & \phantom{0}1.66 & \phantom{0}3.16  &    &    &    &    &      &     &     &     & 2.34 \\ 
\textbf{\caa}    & - & 16.96 & 10.22 & 15.11 & 11.54 & 14.86 &&  & &  &     &     &     &     & 13.74 \\ 
\cdashline{1-16}[1pt/1pt]
\textbf{\ours} & - & \textbf{18.85}\rlap{\textsuperscript{$\dagger$}} & \textbf{10.30}\rlap{\textsuperscript{$\dagger$}} & \textbf{15.24}\rlap{\textsuperscript{$\dagger$}} & \textbf{11.71}\rlap{\textsuperscript{$\dagger$}} & \textbf{15.05}\rlap{\textsuperscript{$\dagger$}} &    &    &    &    &      &     &     &     & \textbf{14.23}\rlap{\textsuperscript{$\dagger$}} \\ \hline
\rowcolor[gray]{.93} \textbf{\mgsm}         & \textbf{en} & \textbf{de} & \textbf{es}    & \textbf{fr}    & \textbf{ja}    & \textbf{ru}    & \textbf{sw}    & \textbf{th}    & \textbf{zh}    &     &     &     &     &     & \\
\textbf{\baseline}    & 51.20       & 46.40   & 42.40   & 42.40   & 35.20   & 38.40   & 34.80   & 35.60   & 39.60   &     &     &     &     &     & 39.35   \\
\textbf{\mtgoogle} & -  & 46.00 &	50.40\rlap{\textsuperscript{$\dagger$}} &	47.20\rlap{\textsuperscript{$\dagger$}} &	44.40\rlap{\textsuperscript{$\dagger$}} &	46.80\rlap{\textsuperscript{$\dagger$}} &	45.60\rlap{\textsuperscript{$\dagger$}} &	45.60\rlap{\textsuperscript{$\dagger$}} &	47.60\rlap{\textsuperscript{$\dagger$}} &	   &     &     &     &     &46.70\rlap{\textsuperscript{$\dagger$}} \\
\textbf{\mtllm}  & -  & 20.40 &  	38.80 &  	32.40 &  	10.80 &  	18.40 &  	22.00	 &  \phantom{0}1.60 &  26.80    &     &     &     &   & 	 & 21.40 \\
\textbf{\iti}         & - & 46.00   & 43.20   & 44.80   & 35.60   & 40.00   & 36.80   & 34.80   & 42.80   &     &     &     &     &     & 40.50   \\ 
\textbf{\caa}    & - &  42.40 & 42.00 & 40.00 & 34.40 & 40.80 & 36.20 & 34.40 & 45.20  &     &     &     &     &     & 39.43 \\ 
\cdashline{1-16}[1pt/1pt]
\textbf{\ours} & - & \textbf{48.40}\rlap{\textsuperscript{$\dagger$}} & \textbf{46.80}  & \textbf{45.20} & \textbf{37.60} & \textbf{44.80}  & \textbf{38.00}  & \textbf{38.80}  & \textbf{43.20} &     &     &     &     &     & \textbf{42.85} \\
\bottomrule
\end{tabular}
\caption{\label{tab:overall} The overall results of nine NLP tasks with multilingual intervention. $\dagger$ denotes the best results.}
\end{table*}

%% file: tables/fr.tex
\begin{table*}[h]\small
\centering
\begin{tabular}{lcccccccccc}
\toprule
& \textbf{en}   & \textbf{ar}   & \textbf{el}   & \textbf{es}   & \textbf{hi}   & \textbf{ru}   & \textbf{tr}   & \textbf{vi}   & \textbf{zh}   & \textbf{AVG}  \\ \midrule
\textbf{\baseline}& 45.11 & 44.70 & 15.37 & 39.37 & 50.18 & 36.99 & 10.51 & 38.77 & 42.20 & 35.91 \\
 \textbf{\ours }& \textbf{52.36} & \textbf{52.33} & \textbf{15.62} & \textbf{51.37} & \textbf{55.40} & \textbf{39.69} & \textbf{10.94} & \textbf{46.48} & \textbf{47.14} &\textbf{ 41.26} \\
\bottomrule
\end{tabular}
\caption{\label{tab:xx2fr} \ours on the Many-to-French translation task.}
\end{table*}